\documentclass[runningheads]{llncs}
\usepackage[T1]{fontenc}
\usepackage{graphicx}
\begin{document}
\title{Arbitrarily Applicable Same/Opposite Relational Responding with NARS}
\author{Robert Johansson\inst{1} \and
Patrick Hammer\inst{1,2} \and
Tony Lofthouse\inst{1}}
\authorrunning{R. Johansson et al.}
\institute{Department of Psychology, Stockholm University, Stockholm, Sweden \and
Division of Robotics, Perception and Learning, KTH Royal Institute of Technology, Stockholm, Sweden\\
\email{\{robert.johansson, patrick.hammer, tony.lofthouse\}@psychology.su.se}}
\maketitle
\begin{abstract}
Same/opposite relational responding, a fundamental aspect of human symbolic cognition, allows the flexible generalization of stimulus relationships based on minimal experience. In this study, we demonstrate the emergence of \textit{arbitrarily applicable} same/opposite relational responding within the Non-Axiomatic Reasoning System (NARS), a computational cognitive architecture designed for adaptive reasoning under uncertainty. Specifically, we extend NARS with an implementation of \textit{acquired relations}, enabling the system to explicitly derive both symmetric (mutual entailment) and novel relational combinations (combinatorial entailment) from minimal explicit training in a contextually controlled matching-to-sample (MTS) procedure. Experimental results show that NARS rapidly internalizes explicitly trained relational rules and robustly demonstrates derived relational generalizations based on arbitrary contextual cues. Importantly, derived relational responding in critical test phases inherently combines both mutual and combinatorial entailments, such as deriving same-relations from multiple explicitly trained opposite-relations. Internal confidence metrics illustrate strong internalization of these relational principles, closely paralleling phenomena observed in human relational learning experiments. Our findings underscore the potential for integrating nuanced relational learning mechanisms inspired by learning psychology into artificial general intelligence frameworks, explicitly highlighting the arbitrary and context-sensitive relational capabilities modeled within NARS.
\end{abstract}

\section{Introduction}

Humans exhibit a remarkable ability to generalize symbolic relationships beyond explicitly trained examples. This capability, explained by Relational Frame Theory (RFT), is termed \textit{Arbitrarily Applicable Relational Responding} (AARR). AARR specifically describes responding to stimuli based on arbitrary contextual cues, rather than intrinsic physical similarities, enabling crucial cognitive skills such as language comprehension, abstract reasoning, and symbolic manipulation~\cite{hayes2001}. AARR involves rapid inference of novel relationships through mutual entailment (symmetry; if stimulus $A$ relates to $B$, then $B$ relates to $A$) and combinatorial entailment (transitivity; from $A$ relates to $B$ and $B$ relates to $C$, infer $A$ relates to $C$). Historically, experimental paradigms such as Matching-to-Sample (MTS) tasks have demonstrated how relational responding spontaneously generalizes beyond direct training~\cite{hayes2001}.

Computationally modeling relational responding presents a significant challenge for Artificial General Intelligence (AGI) because it requires dynamically forming and manipulating relational structures based on minimal explicit training. Successful computational models must generalize relational knowledge flexibly, reflecting the relational sensitivity observed in human cognition. Achieving this capability computationally not only contributes to artificial intelligence development but also deepens our understanding of human symbolic cognition.

In this paper, we propose a computational model of same/opposite relational responding implemented within the Non-Axiomatic Reasoning System (NARS), an adaptive cognitive architecture explicitly designed for reasoning under uncertainty and limited resources~\cite{wang2013nalbook}. We introduce an extension called \textit{acquired relations}, enabling NARS to explicitly derive relational patterns directly from sensorimotor experience. Through structured MTS procedures, we illustrate NARS' capability to explicitly learn and generalize mutual and combinatorial entailments, essential to same/opposite relational responding. To our knowledge, this study provides the first computational demonstration of emergent same/opposite relational responding—explicitly incorporating mutual and combinatorial entailments—within a cognitive architecture such as NARS.

Our findings demonstrate the feasibility of integrating cognitive psychological theories of relational reasoning with computational cognitive architectures, potentially enabling more flexible and human-like symbolic reasoning capabilities within artificial intelligence systems.

\section{Background}

\subsection{Arbitrarily Applicable Relational Responding and Relational Frame Theory}

Relational Frame Theory (RFT) is a behavioral account of human symbolic reasoning, proposing that core cognitive abilities such as language, analogy-making, and abstract thinking emerge from a learned skill called \textit{arbitrarily applicable relational responding} (AARR)~\cite{hayes2001}. AARR refers to the uniquely human capacity to flexibly relate stimuli based on arbitrary contextual cues rather than inherent physical properties. Crucially, stimulus equivalence is considered a specific instance of AARR involving the relational frame of coordination (i.e., sameness).

RFT defines two core relational properties: \textit{mutual entailment} (symmetry) and \textit{combinatorial entailment} (transitivity). Mutual entailment describes spontaneous bi-directional inference; training the relation $A \rightarrow B$ spontaneously yields $B \rightarrow A$. Combinatorial entailment describes deriving novel relations by combining previously learned relations (e.g., given $A \rightarrow B$ and $B \rightarrow C$, one infers $A \rightarrow C$). Importantly, these relational responses depend on contextual control rather than associative strength alone, distinguishing human symbolic reasoning from simpler associative learning observed in non-human animals.

Thus, RFT provides a functional framework for understanding human symbolic cognition, emphasizing relational responding as a generalized, operantly learned behavior underlying complex cognitive phenomena.

\subsection{Matching-to-sample task in RFT}

\begin{figure}[h!]
\centering
\includegraphics[width=\textwidth]{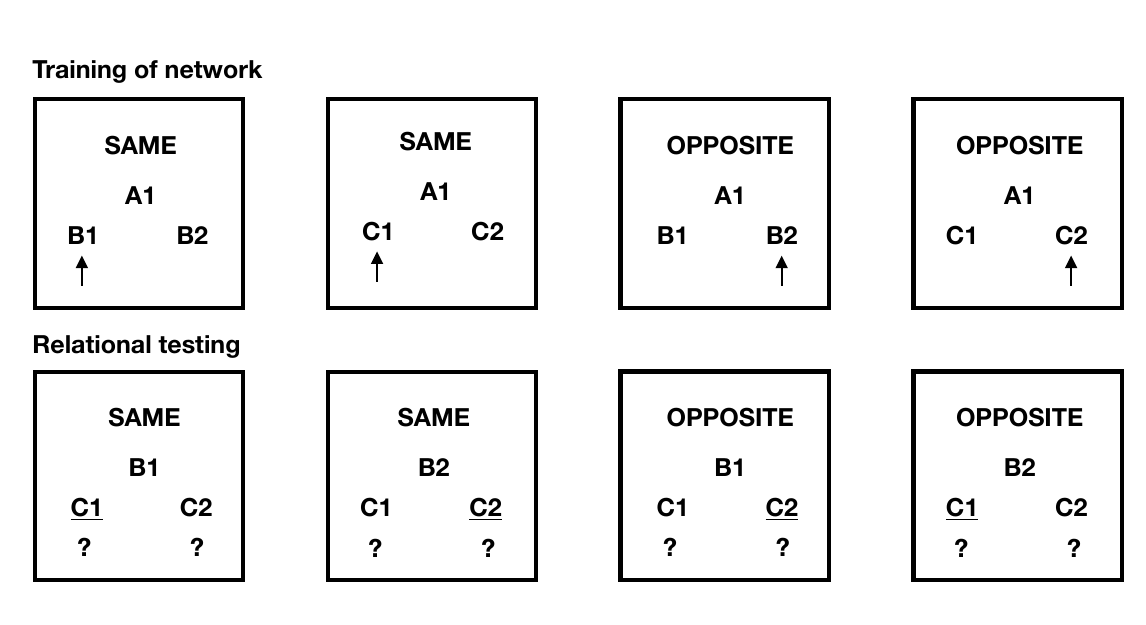}
\caption{The Matching-to-Sample task used in the present study. Pre-training phases are excluded from this figure. The figure illustrates experimental phases 2–3, with underlined options indicating correct responses.}
\label{fig:mts_aarr}
\end{figure}

The Matching-to-Sample (MTS) task is a widely-used experimental paradigm in the study of relational responding, particularly within the framework of Relational Frame Theory (RFT). In a typical MTS procedure, participants are first presented with a sample stimulus and then choose from comparison stimuli based on relational rules explicitly taught or inferred from context. A relational contextual cue (such as SAME or OPPOSITE) can also be presented above the sample stimulus, specifying the relational response required. Successful performance relies on participants’ ability to derive relational responses, illustrating mutual entailment and combinatorial entailment. The MTS paradigm thus effectively assesses participants’ capacity for arbitrarily applicable relational responding, especially regarding stimulus equivalence and relational generalization. Figure~\ref{fig:mts_aarr} illustrates a typical Matching-to-Sample task scenario used in RFT research.

\subsection{Same/Opposite Relational Responding}

Within RFT, relational responding includes diverse relational frames such as sameness, opposite, distinction, comparison, and hierarchy~\cite{hayes2001}. The relational frame of opposition involves responding to stimuli as functionally opposite based on contextual cues. Combining relational frames (e.g., SAME and OPPOSITE) through mutual and combinatorial entailments enables sophisticated cognitive behaviors like categorization, discrimination, and generalized symbolic reasoning.

Explicitly training opposite relations (e.g., $A$ opposite $B$, $A$ opposite $C$) can yield derived sameness relations ($B$ same $C$), demonstrating the combinational logic central to human relational cognition. While computational models explicitly inspired by RFT have primarily addressed stimulus equivalence~\cite{tovar2023computational}, models explicitly incorporating multiple relational frames remain rare (but see \cite{edwards2022functional}). To our knowledge, computational demonstrations explicitly modeling same/opposite relational responding with mutual and combinatorial entailments have yet to be presented.

\subsection{The Non-Axiomatic Reasoning System}

The Non-Axiomatic Reasoning System (NARS) is a computational cognitive architecture designed to model human-like reasoning under uncertainty and limited resources~\cite{wang2013nalbook}. NARS employs Non-Axiomatic Logic (NAL), continuously updating beliefs based on real-time interaction and experience, thus closely mirroring human cognitive flexibility and adaptive reasoning.

NARS naturally supports deriving new relationships from minimal experiential evidence, making it well-suited for modeling relational responding. Prior research has demonstrated NARS’ capability in basic relational tasks, such as generalized identity matching, via adaptive sensorimotor reasoning~\cite{johansson2023stimulus}. The current study explicitly builds upon these findings by introducing acquired relations and systematically examining complex same/opposite relational responding within NARS. This work represents a novel computational demonstration explicitly incorporating mutual and combinatorial entailments, advancing computational modeling toward more human-like symbolic reasoning capabilities.

\section{Implementation in NARS}

In this study, we extended the Non-Axiomatic Reasoning System (NARS) by explicitly introducing \textit{acquired relations}, enabling the system to derive generalized relational structures from direct sensorimotor interactions. This implementation explicitly realizes the theoretical mechanisms proposed in our previous work regarding stimulus equivalence with NARS~\cite{johansson2023stimulus}, extending beyond equivalence to explicitly model diverse relational frames such as same/opposite responding. Below, we explicitly detail how relational responding was implemented and operationalized within NARS, providing representative examples in the formal language (Narsese) for clarity.

\subsection{Same/Opposite Matching-to-Sample Task in Narsese}

The Matching-to-Sample (MTS) procedure was explicitly represented using Narsese temporal statements, capturing relationships among stimuli, locations, and relational context. This type of encoding has been described extensively elsewhere \cite{johansson2024empirical,johansson2024machine}. A representative trial is encoded as follows:

\begin{verbatim}
<(rel * SAME) --> (loc * ocr)>. :|:
<(sample * X1) --> (loc * ocr)>. :|:
<(left * Y1) --> (loc * ocr)>. :|:
<(right * Y2) --> (loc * ocr)>. :|:
G! :|:
\end{verbatim}

This Narsese representation explicitly encodes stimulus-location pairings (e.g., stimulus \texttt{X1} at the location \texttt{sample}), relational context (\texttt{SAME}), and a goal (\texttt{G!}) prompting action selection. Initially, through exploratory sensorimotor interaction ("motor babbling"), NARS spontaneously executes the \texttt{\^{}match} operation, for instance, selecting the left stimulus. Upon correct responding, NARS explicitly derives a corresponding relational contingency:

\begin{verbatim}
(<(rel * SAME) --> (loc * ocr)> &/ 
 <(sample * X1) --> (loc * ocr)> &/
 <(left * Y1) --> (loc * ocr)> &/
 <({SELF} * (sample * left)) --> ^match>) =/> G>.
\end{verbatim}

This explicitly derived contingency captures the relationship between context (\texttt{SAME}), stimulus identity (\texttt{X1}, \texttt{Y1}), and locations (\texttt{sample}, \texttt{left}), forming the foundational relational hypothesis supporting generalized responding.

\subsection{Acquired Relations and Generalization}

Acquired relations explicitly abstract relational patterns from sensorimotor interactions into general relational hypotheses:

\begin{verbatim}
<(X1 * Y1) --> (ocr * ocr)> &&
<(sample * left) --> (loc * loc)>.
\end{verbatim}

These explicitly represent generalized relationships between stimulus identities (e.g., \texttt{X1} and \texttt{Y1}) and stimulus locations (e.g., \texttt{sample} and \texttt{left}). Once established, these acquired relations explicitly generate higher-order implications linking stimulus identities and locations to behavioral outcomes:

\begin{verbatim}
(<(X1 * Y1) --> (ocr * ocr)> &&
 <(sample * left) --> (loc * loc)>) ==> 
    (<(sample * X1) --> (loc * ocr)> &/
     <(left * Y1) --> (loc * ocr)> &/
     <({SELF} * (sample * left)) --> ^match>) =/> G>.
\end{verbatim}

Further generalization involves abstracting to variable placeholders ($1, $2, $3, $4):

\begin{verbatim}
(<($1 * $2) --> (ocr * ocr)> &&
 <($3 * $4) --> (loc * loc)>) ==> 
    (<($3 * $1) --> (loc * ocr)> &/
     <($4 * $2) --> (loc * ocr)> &/
     <({SELF} * ($3 * $4)) --> ^match>) =/> G>.
\end{verbatim}

This generalized relational schema explicitly facilitates reasoning across novel stimulus sets and contexts, significantly enhancing NARS' scalability and flexibility in relational generalization.

\subsection{Explicit Relational Naming}

In addition to implicitly represented relational structures, explicit relational naming was introduced for greater clarity and interpretability. After learning explicit contingencies such as:

\begin{verbatim}
(<(rel * SAME) --> (loc * ocr)> &/ 
 <(sample * X1) --> (loc * ocr)> &/
 <(left * Y1) --> (loc * ocr)> &/
 <({SELF} * (sample * left)) --> ^match>) =/> G>,
\end{verbatim}

NARS explicitly abstracted and internally represented this knowledge as named relational statements:

\begin{verbatim}
<(X1 * Y1) --> SAME>.
\end{verbatim}

This explicitly named relational form can equivalently be represented as:

\begin{verbatim}
<(SAME * (X1 * Y1)) --> (ocr * (ocr * ocr))>.
\end{verbatim}

Such explicitly named relational representations enhance NARS' symbolic clarity, directly supporting novel relational derivations during relational testing phases.

In summary, through explicitly introducing acquired relations and relational naming, we provided a novel computational realization of Arbitrarily Applicable Relational Responding (AARR) within NARS. These theoretical extensions equip NARS with sophisticated relational reasoning capabilities, supporting emergent relational generalization aligned with human-like symbolic reasoning processes.

\section{Experimental Setup}

The experiment was structured into three phases, explicitly designed to systematically establish and evaluate same/opposite relational responding within the Non-Axiomatic Reasoning System (NARS). Each phase is detailed below, clearly articulating how mutual and combinatorial entailments were operationalized and evaluated using the Matching-to-Sample (MTS) procedure (see Figure~\ref{fig:mts_aarr}). All phases included four blocks of 16 trials each.

\begin{figure}[h!]
\centering
\includegraphics[scale=0.10]{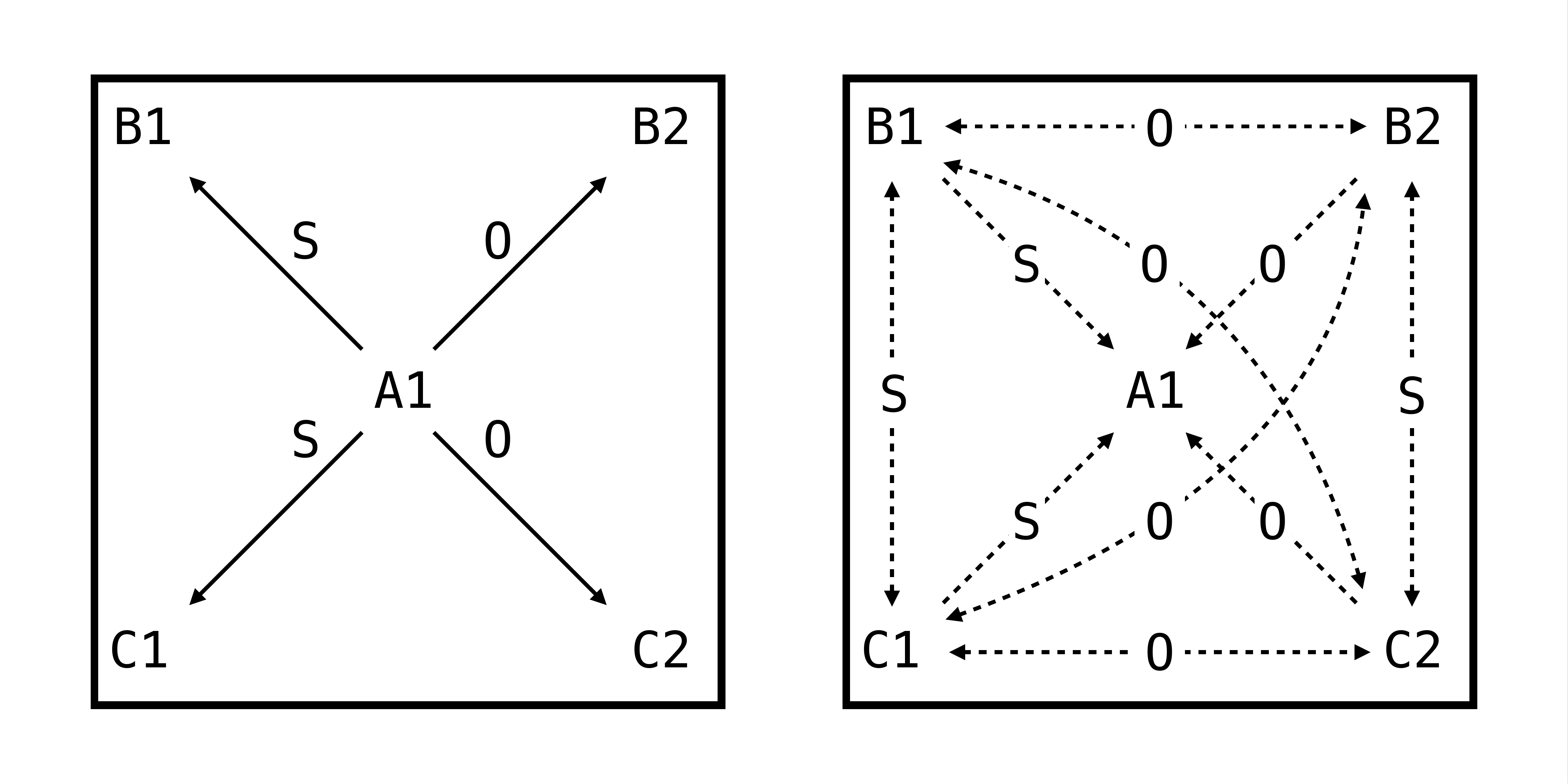}
\caption{Explicitly trained (left panel) and derived (right panel) relational networks from the present study. \texttt{S} and \texttt{O} denote SAME and OPPOSITE, respectively.}
\label{fig:derived_relations}
\end{figure}

\paragraph{Phase 1: Explicit Pretraining of Relational Frames (Mutual and Combinatorial Entailment).}

Foundational relational responding capabilities were explicitly trained prior to evaluating emergent relational responding. Specifically, NARS was trained explicitly on symmetric (mutual entailment) and transitive (combinatorial entailment) relational frames using SAME and OPPOSITE contexts. Mutual entailment involved explicit training of symmetrical relations (if $X \rightarrow Y$, explicitly train $Y \rightarrow X$), ensuring NARS derived symmetrical relations spontaneously. Combinatorial entailment involved explicit training of transitive relations (if $X \rightarrow Y$ and $Y \rightarrow Z$, explicitly train $X \rightarrow Z$), thus preparing NARS to derive novel relational inferences.

\paragraph{Phase 2: Relational Network Training Using Matching-to-Sample (MTS).}

In Phase 2, NARS underwent relational network training with novel stimulus sets (AB, AC), using the Matching-to-Sample (MTS) paradigm. Stimulus sets AB and AC were arbitrarily selected to ensure relational responding was driven explicitly by contextual cues rather than intrinsic stimulus properties. Stimulus pairs were explicitly reinforced under SAME and OPPOSITE contexts, with correct relational selections consistently receiving positive feedback. Through these sensorimotor interactions, NARS formed internal acquired relations and strengthened relational hypotheses. Figure~\ref{fig:derived_relations} (left-hand side) visually illustrates the explicitly trained relational networks from this phase.

\paragraph{Phase 3: Testing for Emergent Same/Opposite Relational Responding.}

The final testing phase explicitly assessed derived relational responding without reinforcement. Stimulus pairs never explicitly trained (set BC) were tested, requiring NARS to spontaneously combine mutual and combinatorial entailments (e.g., deriving SAME from pairs originally trained as OPPOSITE to a common stimulus). Critically, evaluating derived relational responding without reinforcement explicitly assessed NARS' capacity for spontaneous generalization based solely on previously acquired relational knowledge. Successful performance demonstrated emergent same/opposite relational responding, confirming NARS’ capability to infer novel relations explicitly via acquired relational mechanisms. Figure~\ref{fig:derived_relations} (right-hand side) explicitly illustrates these derived relational networks.

\section{Results}

Figure~\ref{fig:results} summarizes NARS' performance accuracy (\% correct) and internal hypothesis confidence (mutual and combinatorial entailment) across all experimental phases.

\subsection{Explicit Pretraining of SAME and OPPOSITE Relations (Phases XY, YX, YZ, XZ)}

\begin{figure}[h!]
\centering
\includegraphics[width=\textwidth]{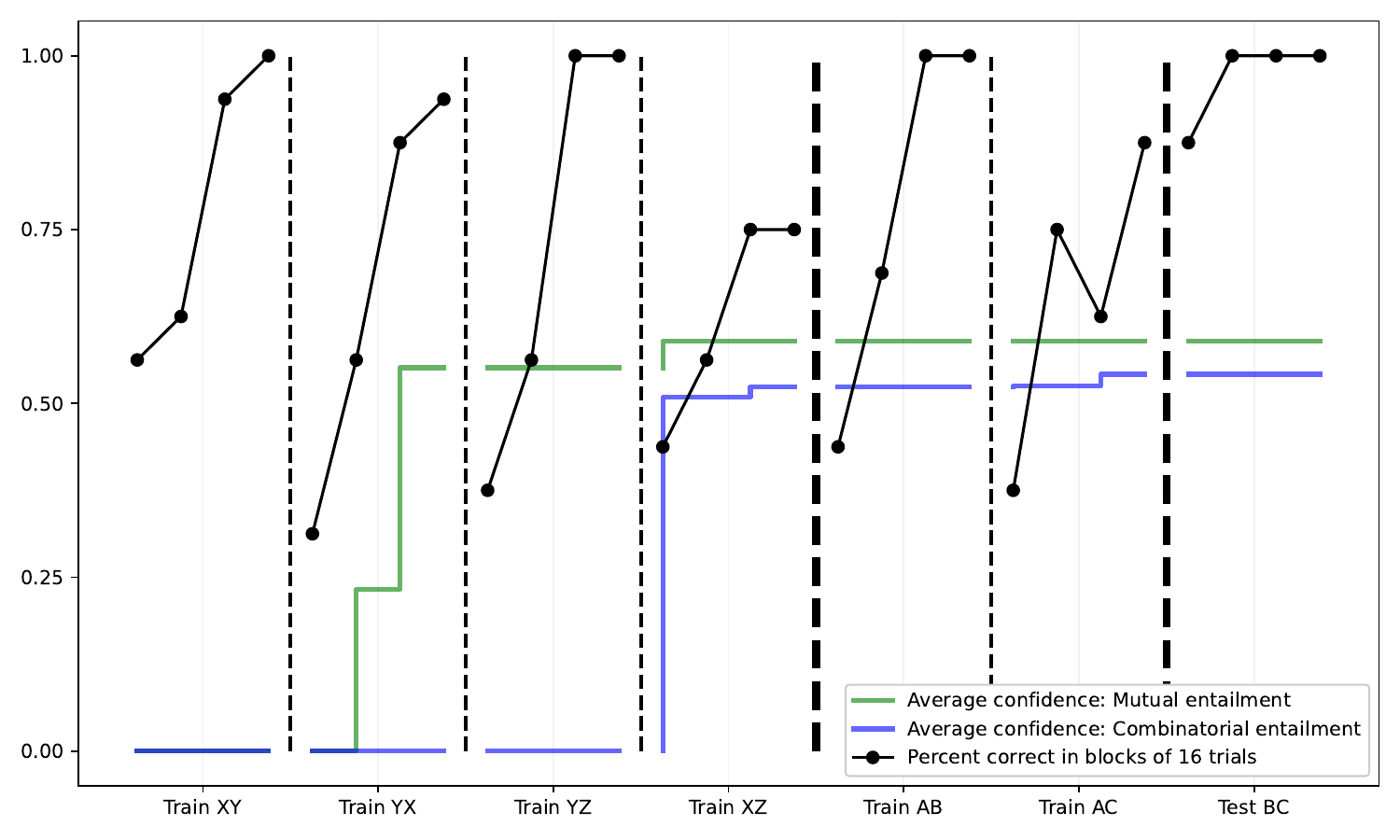}
\caption{Accuracy (\% correct) and internal confidence (mutual entailment and combinatorial entailment) across three experimental phases. Phase 1: Explicit pretraining (XY, YX, YZ, XZ). Phase 2: Relational network training (AB, AC). Phase 3: Derived relational testing (BC). Accuracy is reported per block of 16 trials.}
\label{fig:results}
\end{figure}

NARS rapidly acquired explicitly reinforced SAME and OPPOSITE relations, consistently achieving perfect accuracy (100\%) from early training stages (see Figure~\ref{fig:results}). Concurrently, internal confidence—representing NARS' internally computed certainty levels regarding derived relational hypotheses—increased significantly. Mutual entailment confidence demonstrated rapid, steady growth, explicitly indicating effective internalization of symmetrical relational frames (e.g., deriving \textit{Y SAME X} from explicitly trained \textit{X SAME Y}). By the end of the pretraining phases, mutual entailment confidence approached ceiling levels.

Similarly, combinatorial entailment confidence notably increased, reflecting successful internalization of transitive relational frames (e.g., deriving \textit{X→Z} from explicitly trained \textit{X→Y} and \textit{Y→Z}). Combinatorial entailment confidence stabilized at high levels by the completion of phase XZ, providing robust internal support for subsequent relational learning.

\subsection{Relational Network Training (Phases AB, AC)}

During relational network training with novel stimulus sets (AB, AC), NARS consistently exhibited perfect accuracy (100\%), explicitly demonstrating effective transfer of relational rules established during pretraining. Internal confidence remained high and stable, explicitly confirming that previously generalized relational principles facilitated rapid, accurate acquisition of explicitly trained relational pairs.

\subsection{Derived Relational Testing (Phase BC)}

In the critical derived relational testing phase (BC), involving stimulus pairs never explicitly trained, NARS exhibited perfect accuracy (100\%), explicitly surpassing chance performance (50\%). Surpassing chance explicitly confirms that relational responding was driven by previously internalized relational structures rather than random or associative processes, providing compelling evidence for generalized SAME and OPPOSITE relational responding aligned with Arbitrarily Applicable Relational Responding (AARR) principles.

Internal confidence remained robust for both mutual and combinatorial entailments throughout this derived testing phase, further validating NARS' internal generalization of relational frames, underpinning the flawless observed behavioral accuracy.

\subsection{Example of Internal Relational Representation}

To explicitly illustrate NARS' internal representation of relational rules, we present an exemplar combinatorial entailment hypothesis derived during explicit pretraining:

\begin{verbatim}
<(<($1 * #1) –> SAME> && <(#1 * $2) –> OPPOSITE>) ==>
<($1 * $2) –> OPPOSITE>>
\end{verbatim}

Such internally represented hypotheses explicitly enable NARS to generalize accurately to derived relational tasks, clearly demonstrated during the derived BC testing phase.

In summary, NARS exhibited robust, reliable relational learning across all experimental phases, characterized explicitly by consistently perfect performance and strong internal confidence in mutual and combinational relational frames. Critically, NARS successfully generalized learned relational frames explicitly to novel stimulus pairs without explicit training or feedback, representing a significant advancement in computational modeling of human-like relational responding within the NARS architecture.

\section{Discussion}

The current study demonstrated the emergence of same/opposite relational responding within the Non-Axiomatic Reasoning System (NARS), explicitly showing how mutual and combinatorial entailments combine to generate sophisticated symbolic generalizations. Our findings highlight the feasibility of computationally modeling human-like symbolic reasoning processes described by Relational Frame Theory (RFT), contributing explicitly to a deeper interdisciplinary understanding linking cognitive psychology and Artificial General Intelligence (AGI).

By explicitly introducing acquired relations and relational naming into NARS, we successfully operationalized key RFT principles such as mutual entailment (symmetry) and combinatorial entailment (transitivity). Unlike previous computational models primarily focused on equivalence relations, our approach explicitly demonstrates the computational generalization of multiple relational frames, such as SAME and OPPOSITE. This computational demonstration explicitly underscores the potential of NARS to replicate and generalize relational patterns through minimal training, closely paralleling human relational cognition. Such capabilities are explicitly critical for AGI, suggesting clear pathways toward more flexible, context-sensitive symbolic reasoning in artificial systems. The demonstrated relational flexibility could explicitly enhance AGI systems in tasks  involving natural language understanding, complex categorization, or adaptive decision-making in dynamic environments, where nuanced relational reasoning and symbolic generalization are essential.

However, this study has limitations. The relational structures explored were relatively simple and abstract. Future work should explicitly examine more complex relational networks and multiple relational frames, potentially integrating real-world sensorimotor data to enhance ecological validity and applicability in naturalistic contexts. Additionally, addressing scalability and generalizability explicitly remains critical for practical AGI applications, where relational reasoning must operate robustly amidst noisy, ambiguous, or incomplete real-world data.

A further limitation is the reliance on empirical validation rather than formal proofs of completeness or consistency of the acquired-relations mechanism. While the findings empirically demonstrate robust relational responding, formal analyses of theoretical expressiveness or soundness were beyond the study’s scope. Future work could explicitly investigate these formal properties to enhance theoretical rigor.

In conclusion, the present findings explicitly represent a meaningful advancement in computational cognitive modeling, illustrating concretely how psychological theories such as RFT can explicitly inform and enrich AGI architectures. This integration explicitly advances AGI methodologies and clearly illustrates how arbitrary relational contexts underpin human symbolic reasoning, contributing meaningfully toward more robust, flexible, and human-like AI.


\section*{Disclosure Statement}
The authors declare no competing interests.

\bibliographystyle{splncs04}
\bibliography{ref}

\end{document}